\documentclass[letterpaper]{article}
\usepackage{aaai}
\usepackage{times}
\usepackage{helvet}
\usepackage{courier}
\usepackage{multirow}
\usepackage{color}

\usepackage{amsmath}
\usepackage{hyperref}
\emergencystretch=\maxdimen
\usepackage{graphicx}
\usepackage{xcolor}

\begin{document}
%
\title{A Rich Recipe Representation as Plan to Support Expressive Multi-Modal  Queries on Recipe Content and Preparation Process}


\author{
Vishal Pallagani   \And 
Priyadharsini Ramamurthy* \And 
Vedant Khandelwal \And 
Revathy Venkataramanan \AND 
Kausik Lakkaraju \And 
Sathyanarayanan N. Aakur*  \And
Biplav Srivastava \AND
\small 
AI Institute, University of South Carolina  \and
*Oklahoma State University \\
\{vishalp@mailbox.,vedant@mailbox.,revathy@email.,kausik@email.,biplav.s@\}sc.edu\\
    \{pramamu@ostatemail.,saakurn@\}okstate.edu
}

\maketitle
\begin{abstract}
Food is not only a basic human necessity but also a key factor driving a society’s health and economic well-being. As a result, the cooking domain is  a popular use-case to demonstrate decision-support  (AI) capabilities in service of benefits like precision health with tools ranging from information retrieval interfaces to task-oriented chatbots. An AI here should understand  concepts in the food domain (e.g., recipes, ingredients), be  tolerant to failures encountered while cooking (e.g., browning of butter), handle allergy-based substitutions, and work with multiple data modalities (e.g. text and images). However, the recipes today are handled as textual documents which makes it difficult for machines to read, reason and handle ambiguity. This demands a need for better representation of the recipes, overcoming the ambiguity and sparseness that exists in the current textual documents. In this paper, we discuss the construction of a machine-understandable rich recipe representation (R3), in the form of plans, from the recipes available in natural language. R3 is infused with additional knowledge such as information about allergens and images of ingredients, possible failures and tips for each atomic cooking step. To show the benefits of R3, we also present TREAT, a tool for recipe retrieval which uses R3 to perform multi-modal reasoning on the recipe's content  (plan objects - ingredients and cooking tools), food preparation process  (plan actions and time), and media type (image, text). R3 leads to improved retrieval efficiency and new capabilities that were  hither-to not possible in textual representation.
\end{abstract}

\section{Introduction}

Food is an innate psychological need for human life, along with water, warmth and rest. In addition to being a basic human necessity, food is also a key factor driving a society's health and economic well-being. However, maintaining a healthy diet is essential for good health and nutrition as it prevents many chronic non-communicable diseases. The importance of diet and food has attracted increasing attention, leading to the coining of a new term called as \textit{precision nutrition} \cite{chatelan2019precision,wang2018precision} - which seeks to better the health of a person through precise dietary intake based on unique characteristics of an individual such as DNA, race, gender, and lifestyle habits. The process of cooking is a key enabler for precision nutrition, which has led to extensive research in building and deploying decision-support artificial intelligence (AI) tools - varying from information retrieval (IR) using interfaces to task-oriented chatbots \cite{min2019survey,jiang2020food}.

Recent advancements in learning-based approaches have seen many AI researchers channeling their effort in building large datasets and learning vector representations for recipes in the latent space. The learnt vector representations are used for downstream tasks \cite{li2020reciptor,lien2020recipe} such as cooking (ingredient/recipe/cuisine) recommendation \cite{song2022self,tian2022recipe,kim2021recipebowl}, food conversation chatbots \cite{frummet2022can}, and cooking activity recognition \cite{ramadan2022action}.
The learning-based approaches are heavily dependent on the data they are trained on which, for the cooking domain, is the text-based recipes. However, text is a difficult representation for machines, using natural language processing (NLP), to read, reason and handle ambiguity. While learning an embedding - the state-of-the-art NLP method to pre-process documents -  for the cooking domain, there is a need for representation that captures the cooking actions explicitly. Additionally, the recipe embeddings being learned only from the cooking instructions often lack constraints and attributes such as possible allergens, failures, and relation between multi-modal data points. 

In this paper, inspired by PDDL plans, we consider recipes as plans that are composed of a sequence of steps. A step corresponds to a single food preparation action indicated by a verb and instantiated by world objects that are  subject and object of the action (parameters). 
The actions proposed in R3 have additional knowledge such as allergen information, possible failures that can happen during preparation and their workarounds. 
As a preliminary implementation, we have created 25 egg-based recipes in R3 manually from original recipes taken from the RecipeQA dataset \cite{yagcioglu2018recipeqa}. We further created a tool - TREAT, to demonstrate the advantage of using R3 over natural language from the dataset for the task of information retrieval. We demonstrate how TREAT can perform multi-modal constrained queries from the user for recipe retrieval. Constrained queries consist of restrictions specified by the user whilst obtaining information. Two kinds of constraints can be performed by TREAT, namely, outcome (e.g., \textit{I want a recipe without leaf allergy.}) and process (e.g., \textit{Suggest me a recipe that is not fried.}) constraints. 

We make  following  contributions in the paper: (a)  introduce a machine-readable rich recipe representation for recipes inspired by the structure of PDDL plans called R3, (b)  implement the TREAT multi-modal recipe retrieval system which responds to user queries by reasoning on recipe content and process using R3, and (c) evaluate the impact of R3 by comparing recipes in R3 over text representation for the task of (recipe) information retrieval. In the remainder of the paper, we give a background of related work, followed by our approach describing the construction of R3 and its use in the TREAT system. We the demonstrate the effectiveness of using R3 by performing a comparative study and conclude with pointers to future work.

\section{Related Work}

In this section, we describe existing work on representing recipes and plans to motivate the need and benefits of our approach.

\subsection{Representing Recipes}

Building representations for recipes using various learning-based techniques has been active research area in the recent past \cite{li2020reciptor}. However, there are inherently many limitations of learnt recipe embeddings from text. The closest effort to ours in obtaining a new representation for recipes has been to manually convert them to a flowgraph \cite{mori2014flow}. However, R3 differs from flowgraphs, as the former is created with the goal of offering flexibility to be used by both learning and planning based approaches. An automated planner cannot work with the representation from a flowgraph, but can do so on R3 enabling additional functionalities such as generating multiple similar plans (i.e.,similar recipes) or alternative plans in case of failures (e.g., missing ingredient or cooking tool). 

Existing datasets in the cooking domain mostly constitute of text and images \cite{lien2020recipe,leroux2002flavour,yagcioglu2018recipeqa}. There also have been efforts in building a cooking workflow dataset \cite{pan2020multi}. Researchers also have looked at representing recipes in the form of a knowledge graph \cite{shirai2021identifying} but as the graph grows in size, deploying and performing real-time decision support becomes a challenge \cite{haussmann2019foodkg}. R3 has been developed to foster the continual search for finding a better representation of procedural texts and to demonstrate its viability, we adopted the task of recipe retrieval as it is an accepted baseline to validate a new representation \cite{lien2020recipe,chen2016deep,zhu2019r2gan}. Roither et al \cite{roither2022chef} have worked on classifying ingredients into allergen categories. In our work, through our representations, we enable searching recipes without an allergen. This enhances user convenience and ease-of-use.

\subsection{Plan Representation}

Planning is the problem of finding a path to lead an agent from an initial state to a goal state, given a set of legal actions to transition between states. 
The field of planning has seen many representations. In classical planning, which assumes complete information about the world, there was STRIPS~\cite{strips}, Action Description Language (ADL)~\cite{adl} and SAS+~\cite{comparison-pl-backstorm} before Planning Domain Description Language (PDDL)~\cite{pddl,pddl2.1} standardized the notations. 

In PDDL, a planning  environment is described in terms of objects in  the world, predicates that describe relations that hold between these objects, and actions that bring change to the world by manipulating relations. 
To faciliate automated solving, PDDL envisages two files, a domain description file which specifies information independent of a problem like predicates and actions, and a problem description file which specifies the initial and goal states. PDDL provides a relaxed specification of the output plan --  it is a series of time steps, each of which can have one or more  instantiated actions with concurrency semantics. Specification of the correctness of plans was formalized much later when tools like VAL were created~\cite{plan-verif-val} but they assume access to domain and problem files.

There is a long history in planning of viewing plans independently from how they may have been generated or evolve over time  e.g., by an automated planner or by a human. In \cite{myers-plangen}, the need for diverse planning was motivated and grounded in a meta-theory allowing the plans to be modified and compared. In 
\cite{plan-lifecycle}, the authors describe a platform to store, search and manage plans, representing Information Technology workflows, as they are evolve over time.

In this work, we adopt the plan representation of PDDL for recipes but do not assume how recipes may have been created. We thus allow not only efficient management of recipe data (storage and retrieval) but also enable future automation in recipe creation (by reuse and composition of recipes to create new recipes), and food preparation (by monitoring the execution of agents) by allowing reasoning with both the data and control flow inherent in the plan-based  recipe representation consisting of text and images.

\section{Approach}

\begin{figure*}[!ht]
	\centering
	\includegraphics[width=0.65\textwidth]{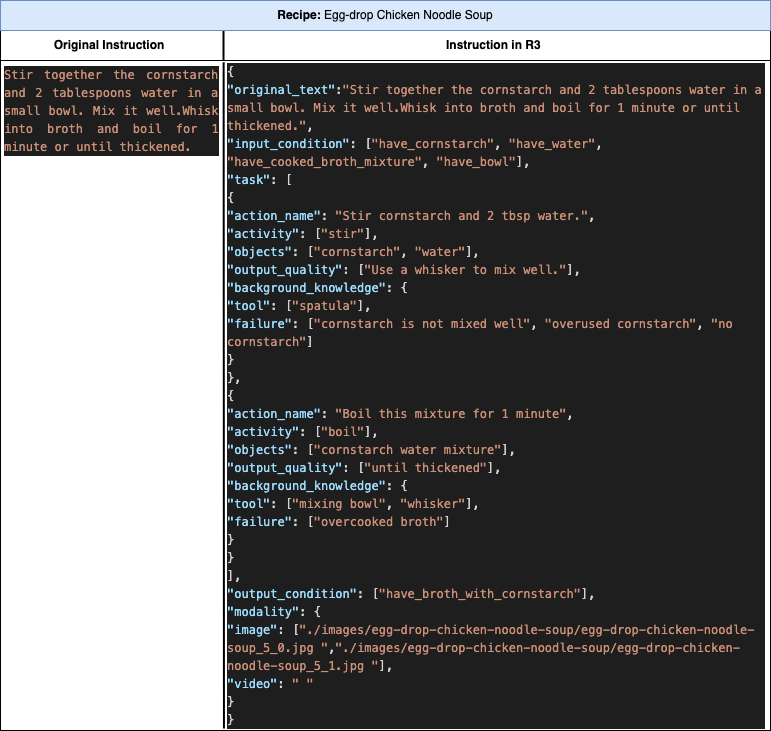}
	\caption{Difference between textual representation and R3 for a single instruction}
	\label{fig:representation_comparison}
\end{figure*}

In this section, we first describe out plan-based recipe representation (R3) and then describe a system, TREAT, that uses it to support information retrieval. 


\begin{figure*}[!ht]
	\centering
	\includegraphics[width=0.7\textwidth]{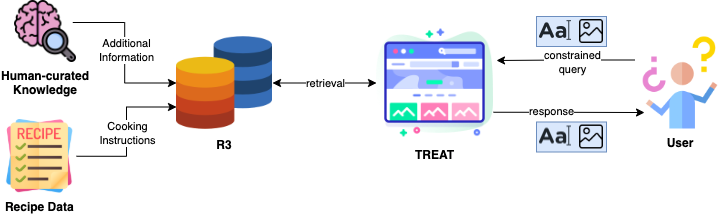}
	\caption{Data flow between TREAT and R3}
	\label{fig:sysarch}
\end{figure*}



\begin{figure}[!ht]
	\centering
	\includegraphics[width=0.3\textwidth]{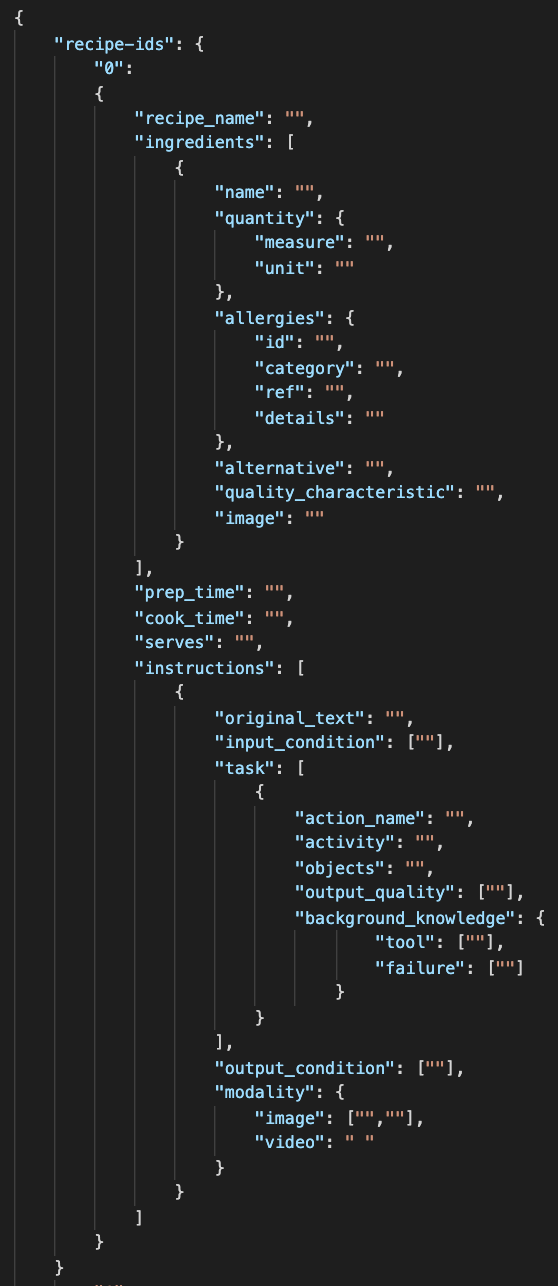}
	\caption{Skeletal structure of R3.}
	\label{fig:represen}
\end{figure}

\begin{figure}[!b]
	\centering
	\includegraphics[width=0.4\textwidth]{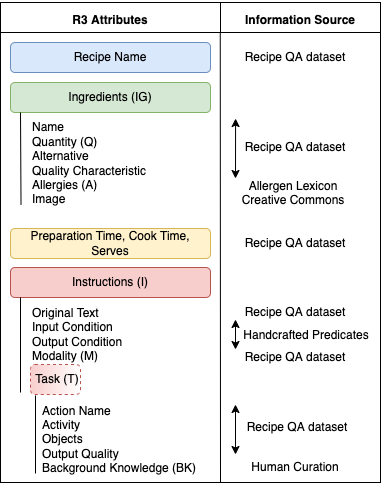}
	\caption{Information source for different attributes in \textbf{R3}.}
	\label{fig:infosource}
\end{figure}

\subsection{Rich Recipe Representation}


We define a recipe \textbf{(R)} as a tuple consisting of a required set of ingredients \textbf{(IG)}, preparation time, cooking time, number of people served, and is said to be complete on performing a series of instructions \textbf{(I)}. Each ingredient is a set consisting of the name, quantity \textbf{(Q)} required for the recipe, allergen information \textbf{(A)}, alternative replacements for the ingredient, quality characteristic which signifies on state of the ingredient (e.g. grated cheese or cheese slice are two different quality characteristics for cheese) and corresponding image. \textbf{Q} is a tuple consisting of measurement and corresponding units. \textbf{A} captures the allergen information - ID to index all possible allergens, category of allergen to which the ingredient belongs, reference to the source from where this information is captured, and additional details for the allergen in the Knowledge Graph.

An instruction is broken down into atomic actions. \textbf{I} is a tuple consisting of the original recipe text from which the representation is being built, the input condition and output condition which define the requirements and changes that happen in the cooking environment once an instruction is performed, tasks \textbf{(T)} which captures the atomic actions and modality \textbf{(M)} which has visual information regarding the instruction. The set \textbf{T} also contains information regarding the objects used and the activities being performed on them, output quality of the action, and background knowledge \textbf{(BK)}. \textbf{BK} is a 2-tuple consisting of tools and failures associated with the atomic action. Figure \ref{fig:represen} shows the skeletal structure of the R3. In an initial effort, we have manually created R3 for 25 egg-based recipes obtained from the RecipeQA dataset. We show the difference between the original representation and R3 for a single instruction of \textit{Egg-drop Chicken 
Noodle Soup} in Figure \ref{fig:representation_comparison}. We now proceed to explain how we created a proof-of-concept corpus of recipes in R3 and   how this representation is vital in building a prototypical recipe retrieval system - TREAT.


\subsection{Populating R3} 
Most of the attributes in R3 are analogous to the data points present in RecipeQA. However, there are additional attributes which are modeled using manual curation of relevant data from Creative Commons. Figure \ref{fig:infosource} gives an overview of how each attribute is populated in R3, out of which, \textit{allergies} needs more explanation. 

\begin{figure}[!ht]
	\centering
	\includegraphics[width=0.4\textwidth]{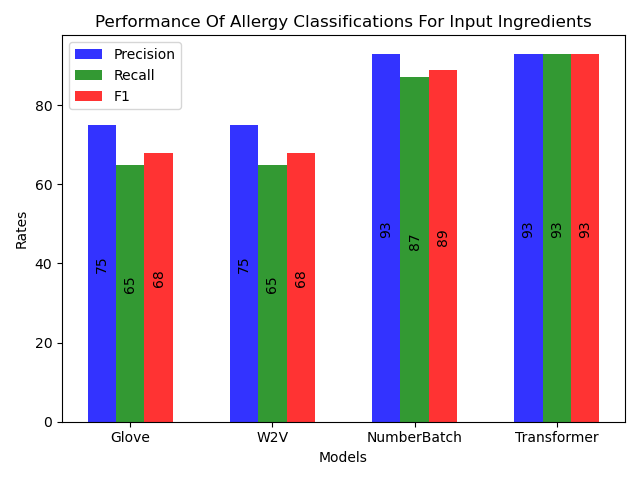}
	\caption{Comparison of different word embedding models for ingredient similarity.}
	\label{fig:ing_models}
\end{figure}

For a given recipe, it is the ingredient(s) to which a consumer is allergic to and this would also help in making intelligent AI models. For instance, a current AI system might classify \textit{pizza} as a dish containing \textit{gluten} allergy, but that pizza might be completely made up of gluten-free ingredients. Thus, in R3 we focus on the possibility of allergies that can be faced by a consumer based on the ingredients. We have created an allergen lexicon with 17 different classes following the guidelines provided by the Institute of Agriculture and Natural Resources \footnote{\url{https://ianr.unl.edu/}}.

However, all the ingredients that belong to a particular allergen class are not present in the lexicon. Thus, we performed a comparative study of multiple word embedding methods (Word2Vec \cite{mikolov2013efficient}/Conceptnet Numberbatch \cite{speer2017conceptnet}/GloVe \cite{pennington2014glove}/Transformers \cite{vaswani2017attention}) to find similar ingredients to one's already present in the allergen lexicon. Figure \ref{fig:ing_models} summarizes the performance of the mentioned models. We see that Transformer outperforms the others in finding similar ingredients. For example, (Transformer) embeddings help us establish the similarity between an unseen ingredient such as \textit{yolk} or \textit{egg whites} with egg which is present in the allergen lexicon. This helps us in deriving the allergy information to unseen ingredients present in R3.

\subsection{TREAT for Recipe Retrieval}

R3 can be used in building decision-support tools in the food domain, out of which information retrieval is considered in this paper. We build a simple web-based tool called TREAT, which makes use of R3 to understand and answer user's constrained queries - which can be either text, image or a combination of both. Figure~\ref{fig:sysarch} shows the overall data flow happening between R3 and the TREAT system based on the user query.

\begin{figure}[!b]
	\centering
	\includegraphics[width=0.45\textwidth, height=0.25\textwidth ]{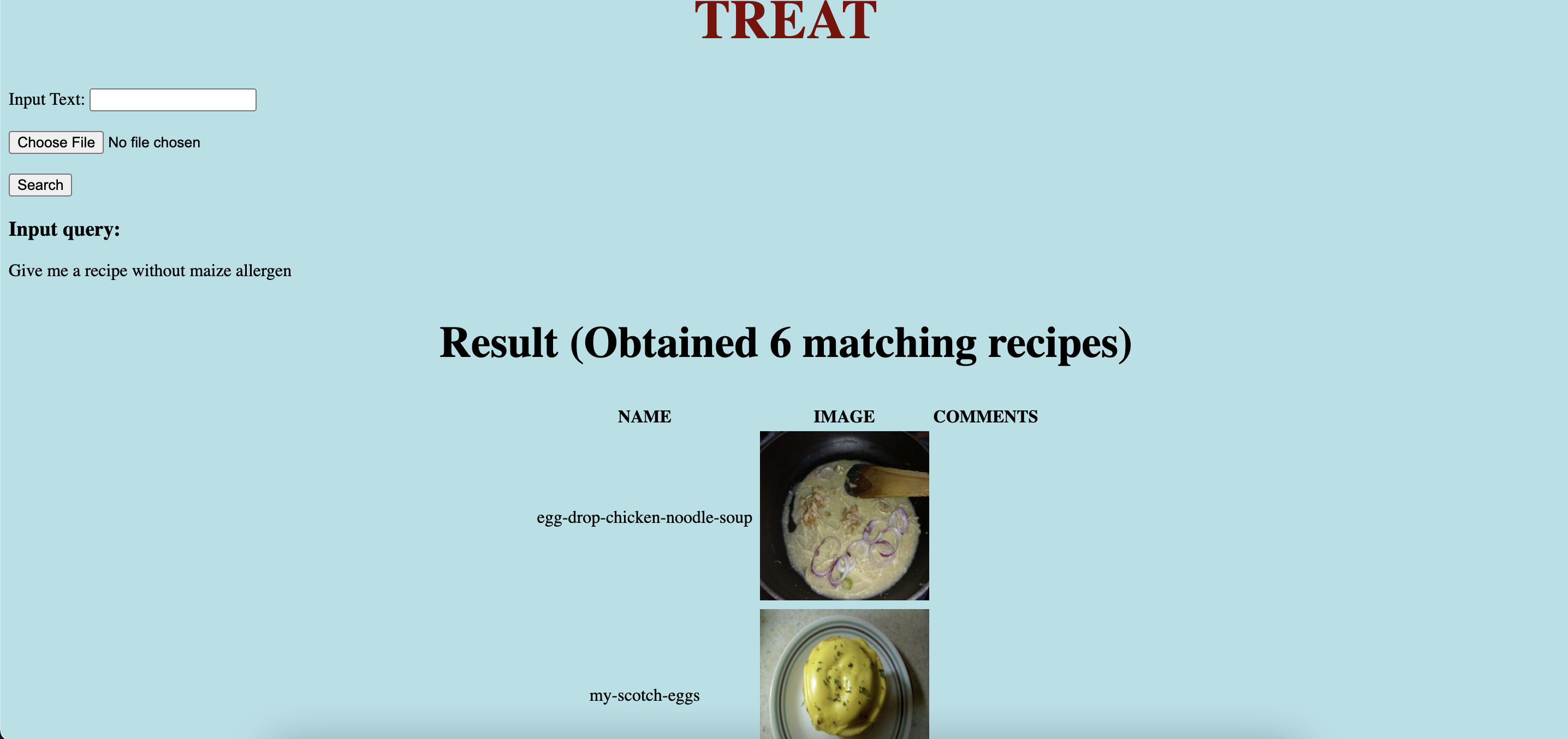}
	\caption{User Interface of TREAT, showing a query where the user is searching for recipes without maize allergen}
	\label{fig:demo}
\end{figure}

\begin{figure}[!b]
	\centering
	\includegraphics[width=0.45\textwidth, height=0.25\textwidth ]{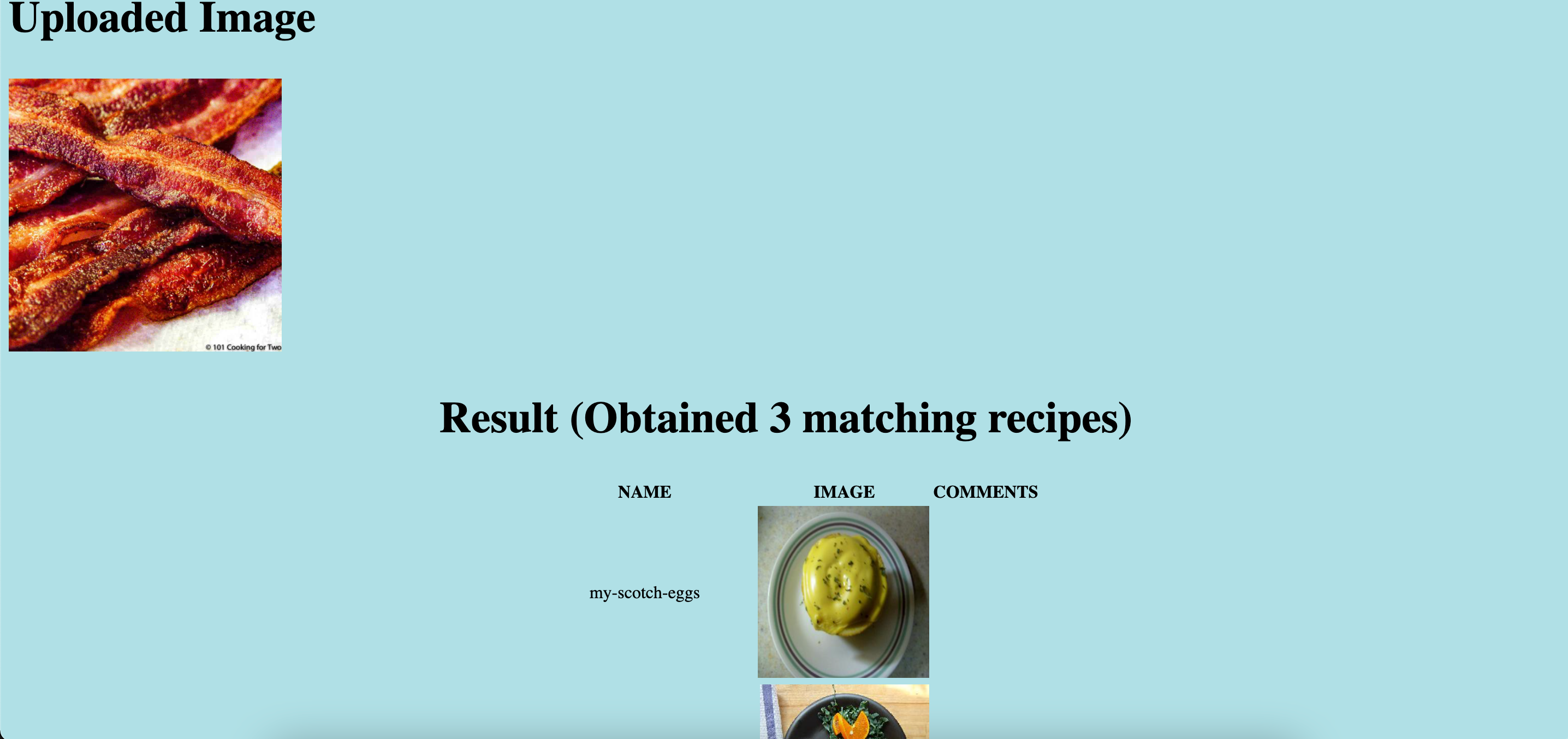}
	\caption{An image query using TREAT. The query is an image of bacon asking for recipes containing it.}
	\label{fig:demo2}
\end{figure}

A user can query TREAT using multiple modalities (text, image, text + image) in order to perform constrained queries, which are of two types:
\begin{itemize}
    \item \textbf{Process Constraints:} where the user puts restrictions on the cooking process, e.g., \textit{Give me a recipe with less than 5 steps and is completed in 20 minutes} or \textit{Suggest me a recipe which is not deep-fried.} For now, TREAT is equipped with the functionality to filter recipes only by length and time.
    \item \textbf{Outcome Constraints:} where the limitations set by the user bring a change about in the general course of the recipe. For example, if the user asks for the recipe of a cake but sets a constraint that he is allergic to gluten, then the system is expected to retrieve gluten-free cake recipes. TREAT can answer user's queries with limitations set on allergens, ingredients, and cuisine. 
\end{itemize}

When the input is a string, the similarity metric employed to match with attributes in R3 is Levenshtein distance, where the threshold for matching is a hyperparameter which is set to 0.7 by default. For image as an input, which can be of an ingredient or final cooked recipe, similarity is computed using the Scale Invariant Feature Transform (SIFT) algorithm \cite{lowe2004distinctive}. The algorithm locates local features in the images called key points and transforms them to vectors called descriptors. The similarity between two images is obtained by computing the euclidean distance between these descriptors. The similarity scores in both approaches are normalized and set to the same default value. 

Figure \ref{fig:demo} shows a demo query posed to TREAT where the user is seen performing a query with allergen constraint on the recipes - \textit{Give me a recipe without maize allergen?} and the \textbf{TREAT} systems retrieves all 6 relevant recipes. Figure \ref{fig:demo2} shows an image query where an image of bacon is given as input to our system, it retrieves 3 recipes with bacon as an ingredient.

In this section, we presented the R3 representation and a tool which uses it for recipe retrieval. In the next section, we present a preliminary evaluation of TREAT to assess R3's impact.

\section{Experiments and Results}

\begin{table}[!b]
\begin{center}
\begin{tabular}{|l|l|l|l|}
    \hline
    \textbf{Query Type} & \textbf{Based on} & \textbf{CVG} & \textbf{IOU} \\
    \hline 
    Process Constraint & Length & 1.00 & 1.00 \\ 
    \hline
    \multirow{4}{*}{Output Constraint}& Ingredient &1.00&1.00\\
    &Allergen & 0.9  & 0.94\\
    &Name & 1.00 & 1.00\\
    &Image & 0.88 & 0.78\\
    \hline
\end{tabular}
\end{center}
\caption{Results of TREAT System}
\label{tab:ProposedResults}
\end{table}

We want to establish that the representation is helpful in driving better recipes for users needs. We perform an evaluation of the TREAT system using 50 unique queries, consisting of both process and outcome constraints. The queries are generated based on two random functions - first selecting the parameters (such as allergen category, image, recipe name, ingredient) and the other selecting the value for those parameters (such as egg, maize category, scotch-eggs, parsley). This systematic process of generating queries avoids the risk of cherry-picking, and allows us to test our system on various edge cases.

Further, a manual process is used to obtain the ground truth values for each of the queries present. The ground truth is compared with the retrieved output from the system to calculate the evaluation metrics, which are \textbf{Coverage of Ground Truth (CVG)} and \textbf{Intersection Over Union (IOU)}. 
\begin{itemize}
    \item Coverage of Ground Truth (CVG): 
    is defined as the intersection of ground truth and retrieved recipes over the ground truth recipe  - Equation \ref{eq_01}.
    \begin{equation}
         CVG~=~\frac{C(R_r~\cap~R_{gt})}{C_{gt}}
         \label{eq_01}
    \end{equation} 
    In the above equation, $R_r$, $R_{gt}$ refers to the set of retrieved recipe and ground truth recipes for a given query, respectively. The $C$ represents length of the set of recipes and $\epsilon~R$ (set of real numbers).
       
    \item Intersection over Union (IOU): This is intersection of ground truth and retrieved recipe over the union of both - Equation \ref{eq_02}.
    \begin{equation}
        IOU~=~\frac{C(R_r~\cap~R_{gt})}{C(R_r~\cup~R_{gt})}
        \label{eq_02}
    \end{equation}
    In the above equation, $R_r$, $R_{gt}$ refers to the set of retrieved recipe and ground truth recipes for a given query, respectively. The $C$ represents length of the set of recipes and $\epsilon~R$ (set of real numbers).
\end{itemize}

\subsection{Performance of the TREAT system}

Table \ref{tab:ProposedResults} shows the performance of the TREAT system on the test set of queries. Results relevant to the user's search query are obtained using the similarity function mentioned in the Approach. We can see that TREAT is able to achieve a performance of 100\% on queries pertaining to process constraints, owing the effectiveness of elaborate representation in R3. It can also be seen that the system is able to perform well on the output constraints.

\subsection{Role of R3 in Improving Performance}
Table \ref{tab:FunctionComp} shows the a comparative study of the queries supported by R3 and the RecipeQA dataset (original) from which it is built. R3 acts as an enabler for the user to perform 3 additional queries, namely, allergen, image, and length based queries in addition to the queries supported by the original dataset.

\begin{table}[!htbp]
\begin{center}
\begin{tabular}{|l|l|l|} 
 \hline
\textbf{Query Type} &\textbf{ Original} &\textbf{ Proposed} \\ 
 \hline
Allergen Based & No & Yes \\
 \hline
Ingredient Based & Yes & Yes \\  
 \hline
Text data-modality & Yes & Yes\\
 \hline
Image data-modality & No & Yes \\  
 \hline
Length Based & No & Yes \\  
 \hline
Name Based & Yes & Yes \\  
 \hline
\end{tabular}
\caption{Different types of query supported by original and proposed data representation\label{tab:FunctionComp}}
\end{center}
\end{table}

We further evaluate the performance of the two types of representations using the test set of queries. Table \ref{Table:performance_comp} shows that the TREAT system can perform around 40\% better than the original textual representation in enabling the user find answers for the queries.

\begin{table}[!htbp]
\begin{center}
\begin{tabular}{|l|l|l|} 
 \hline
\textbf{Representation Type} &\textbf{CVG} &\textbf{IOU} \\ 
 \hline
Original & 0.54 & 0.61 \\
 \hline
Proposed & 0.96 &  0.94\\  
 \hline
\end{tabular}
\caption{Performance Comparison of Original and Proposed representations\label{Table:performance_comp}}
\end{center}
\end{table}
    
\subsection{Support for Allergy Based Search} 

R3 is bound to achieve better performance than the original procedural text as it captures information in a machine-readable and understandable manner, similar to a Planning domain. Thus, this enables easier search for user queries. However, R3 can help answer a wider set of queries, out of which the queries related to allergens are of utmost importance. There has been extensive effort in building the allergy lexicon and use it to populate R3. Thus, we wanted to evaluate the performance of the TREAT system solely on allergy-constrained queries as well and also see how R3 would fare against the original recipe data.

We divide our allergy based queries into two categories - (1) where the user would \textit{explicitly} mention the allergen in his query and (2) where the user just asks for a query without a particular allergy causing ingredient, thereby, having an \textit{implicit} allergen information in the query. An example of an explicit and implicit allergen query can be \textit{Give me a recipe without maize allergen} and \textit{Give me a recipe without parsley} respectively. The original recipe representation, as Table \ref{Table:performance_comp} is incapable of performing explicit allergy based queries, but can perform implicit allergy queries to some extent, i.e., when the query consists of matching ingredients. Table \ref{Table:AllergenQuery} shows how R3 is able to outperform any kind of allergy based queries in comparison to the original dataset.

\begin{table}[!htbb]
\begin{center}
\begin{tabular}{|l|p{0.15\textwidth}|l|l|} 
 \hline
\textbf{Query} &\textbf{Representation Type} &\textbf{CVG} & \textbf{IOU}  \\ 
 \hline
 \multirow{2}{*}{Explicit allergen} & Original & 0.00 & 0.00\\
    & Proposed & 1.00  & 1.00\\
 \hline
 \multirow{2}{*}{Implicit allergen} & Original & 0.16 & 0.24\\
    & Proposed & 0.9  & 0.94\\
 \hline
\end{tabular}
\end{center}
\caption{Allergen based query results}
\label{Table:AllergenQuery}
\end{table}
\section{Conclusion and Future Work}

In this work, we have curated a PDDL plan inspired representation for recipes with additional knowledge, i.e. R3 for building intelligent decision-support tools in the food domain. In order to validate the effectiveness of R3, we build a tool for recipe retrieval which can perform multi-modal reasoning. After experimental evaluation, we arrive at a conclusion that R3 leads to an improved performance in the task of information retrieval when compared to the original textual representation of recipes. In this initial phase of research, we have created R3 for 25 recipes using a semi-automated approach. However, the datasets in food domain have around a million recipes. Thus, scaling R3 using fully-automated approaches is our current ongoing effort. Having a large-scale R3 data would enable us to experiment and evaluate R3 on various fronts and the learnings might help the research community in exploring representations for other formats of procedural texts.

Beyond efficient information retrieval, a plan based representation could enable future automation in recipe creation by reusing existing recipes, reasoning on their content and composing, as necuessary, to create new recipes based on characteristics desired by the user. It can also help automation in  food preparation by enabling monitoring the execution of agents by allowing reasoning with both the data and control flow inherent in the plan-based recipe representation consisting of text and images. These are avenues for future work.


\bibliographystyle{aaai}
\bibliography{references/foodbibs.bib}

\end{document}